\documentclass[twocolumn]{article}

\pdfoutput=1

\usepackage{graphicx}
\usepackage{amsmath}
\usepackage{amsfonts}
\usepackage{algorithm}
\usepackage{algorithmic}
\usepackage{float}
\usepackage{booktabs}
\usepackage{tabularx}

\usepackage{adjustbox}
\usepackage{arydshln}
\usepackage{multirow}
\usepackage{subcaption}

\makeatletter
\newenvironment{breakablealgorithm}
{
	\begin{center}
		\refstepcounter{algorithm}
		\hrule height.8pt depth0pt \kern2pt
		\parskip 0pt
		\renewcommand{\caption}[2][\relax]{
			{\raggedright\textbf{\fname@algorithm~\thealgorithm} ##2\par}
			\ifx\relax##1\relax 
			\addcontentsline{loa}{algorithm}{\protect\numberline{\thealgorithm}##2}
			\else 
			\addcontentsline{loa}{algorithm}{\protect\numberline{\thealgorithm}##1}
			\fi
			\kern2pt\hrule\kern2pt
		}
	}
	{
		\kern2pt\hrule\relax
	\end{center}
}
\makeatother

\newtheorem{theorem}{Theorem}

\title{A self-attention-based differentially private tabular GAN with high data utility}
\author{Zijian Li\thanks{School of Computer Science, Fudan University, Shanghai, China} \and Zhihui Wang\thanks{Corresponding author, School of Computer Science, Fudan University, Shanghai, China}}

\date{}

\begin{document}

\maketitle

\begin{abstract} 
	Generative Adversarial Networks (GANs) have become a ubiquitous technology for data generation, with their prowess in image generation being well-established. However, their application in generating tabular data has been less than ideal. Furthermore, attempting to incorporate differential privacy technology into these frameworks has often resulted in a degradation of data utility. To tackle these challenges, this paper introduces DP-SACTGAN, a novel Conditional Generative Adversarial Network (CGAN) framework for differentially private tabular data generation, aiming to surmount these obstacles. Experimental findings demonstrate that DP-SACTGAN not only accurately models the distribution of the original data but also effectively satisfies the requirements of differential privacy.
\end{abstract}

\section{Introduction}

With the advancements in differential privacy technology, concerns surrounding data privacy have been substantially alleviated. However, several high-sensitivity domains continue to grapple with a distinct challenge—the scarcity of data possessed by most corporate entities. Deep learning demands a substantial volume of training data to achieve the desired performance, a requirement that becomes increasingly unattainable in specialized fields. Consider the healthcare industry, for instance, where the intricacies of numerous diseases and the limited global patient population render amassing substantial datasets not only difficult but prohibitively costly.

In this context, the emergence of generative models\cite{burda_importance_2016,li_generative_2015,makhzani_adversarial_2016,mescheder_adversarial_2018,rezende_stochastic_2014} has provided a pivotal breakthrough. These techniques excel in learning the underlying data distributions from a small set of training samples and subsequently sampling from these distributions, thereby generating substantial quantities of synthetic data. Furthermore, the fusion of deep learning with game theory knowledge has propelled the widespread attention to Generative Adversarial Networks (GANs)\cite{goodfellow_generative_2020} and their variants. GANs have left a profound mark on modeling fundamental data distributions, capable of producing notably high-quality "fake samples" that were hitherto unattainable through previous methods\cite{salimans_improved_2016,sadoughi_speech-driven_2021,mogren_c-rnn-gan_2016}.

In recent years, several studies have explored the application of differential privacy in the realm of deep learning. In 2016, Abadi et al. introduced a gradient clipping method that injected noise during the training of deep learning models to achieve differential privacy \cite{abadi_deep_2016}. This pioneering approach opened the doors to integrating differential privacy into the domain of deep learning. Subsequently, many differential privacy-based Generative Adversarial Networks (DP-GANs) adopted the DP-SGD (Differential Privacy Stochastic Gradient Descent) training method for their discriminators \cite{abadi_deep_2016}. Since differential privacy operates post-processing and generators do not require direct access to sensitive data, models guided by discriminators trained using differential privacy also inherit privacy protection.

However, despite the impressive performance of Generative Adversarial Networks (GANs), combining them with differential privacy poses some key challenges. As mentioned earlier, differential privacy necessitates the continual introduction of noise into sensitive data during deep learning training, which runs counter to the original purpose of deep learning. Additionally, the application of differential privacy often brings about training difficulties, slow convergence, and the generated quality of models trained using differential privacy techniques is often subpar.

In the context of table generation tasks, Generative Adversarial Networks (GANs) do not exhibit the same level of excellence as they do in image applications. Particularly in the realm of table data generation, the application of GANs for data generation encounters a significant challenge—how to model a distribution that possesses both continuous and discrete characteristics. If the GAN-based framework for table data generation is not well-crafted, introducing differential privacy technology to it can further diminish data utility.

To address these issues, this paper draws inspiration from the data modeling principles of CTAB-GAN+\cite{zhao_ctab-gan_2022} and proposes a novel Conditional Generative Adversarial Network (CGAN) differential privacy framework for table data generation, referred to as DP-SACTGAN. The aim is to overcome these obstacles. Experimental results demonstrate that DP-SACTGAN not only accurately captures the distribution of the original data but also satisfies the requirements of differential privacy.

\section{Related Work}

In recent years, research on differential privacy data synthesis has been steadily advancing, achieving some breakthroughs in challenging applications. GS-WGAN, designed by Chen et al. \cite{chen_gs-wgan_2021}, uses the Wasserstein distance to reduce gradient sensitivity and employs carefully crafted gradient sanitization mechanisms to ensure the differential privacy of the generator, thereby enhancing the stability of training the entire differential privacy generative network. PATE-GAN \cite{jordon_pate-gan_2018} was designed to synthesize multivariate tabular data without compromising the privacy of training data. However, works like PrivGAN \cite{mukherjee_privgan_2020} have pointed out that PATE-GAN may not be suitable for generating image data. On one hand, PATE-GAN uses the PATE method to achieve differential privacy. On the other hand, it employs a novel approach to train a student discriminator using generated samples, with the teacher labeling them using the PATE method. Consequently, the student model can undergo differential privacy training without the need for public data, and the generator can use this process to improve sample quality.

The work of G-PATE \cite{long_g-pate_2021}, after referencing DP-GAN and PATE-GAN, introduces a different idea. G-PATE replaces the discriminator in GAN with a set of teacher discriminators trained on disjoint subsets of sensitive data. Additionally, it designs a gradient aggregator to collect information from teacher discriminators and combines them in a distinct differential privacy manner. G-PATE does not require any student discriminators, as the teacher discriminators connect directly to the student generator. This approach allows G-PATE to utilize privacy budgets more efficiently and approximate the true data distribution better.

DP-MERF \cite{harder_dp-merf_2021} and PEARL \cite{liew_pearl_2022} are two other generative models that use differential privacy embedded in kernel functions in the feature space. Specifically, DP-MERF employs random feature representations of kernel mean embeddings, while PEARL improves DP-MERF by combining feature functions that enhance the generator's learning capability. Both of these models focus on generating a differential privacy-embedded space, calculating distance measures between synthesized data and real data in this space. Although both methods offer rapid training, their results on high-dimensional data are less satisfactory.

In recent advancements in tabular data research, addressing data imbalance has been a critical concern. CTGAN\cite{xu_modeling_2019} has improved the architecture of Generative Adversarial Networks (GANs) by introducing pattern-specific normalization techniques. Additionally, it has enhanced the training process to tackle data imbalance issues by adopting a per-sample training method.Subsequently, CTAB-GAN\cite{zhao_ctab-gan_2021} has emerged as one of the most accurate generative models for tabular data modeling in recent years. It possesses the capability to model both continuous and discrete numerical data, effectively addressing data imbalance and long-tail numerical value challenges.A year later, CTAB-GAN+\cite{zhao_ctab-gan_2022} further improved the utility of synthetic data in classification and regression domains by introducing downstream losses in conditional GANs. Furthermore, CTAB-GAN+ integrated the DP-SGD differential privacy optimization algorithm to enforce stricter privacy guarantees while generating tabular data.

\section{Preliminaries}
\subsection{Differential Privacy}
Differential Privacy provides a formal framework for quantifying the privacy guarantees of data analysis algorithms. It is defined as follows:

\begin{equation}
	\label{eq:dp_definition}
	\Pr[\mathcal{M}(D) \in S] \leq \exp(\varepsilon) \cdot \Pr[\mathcal{M}(D') \in S] + \delta
\end{equation}

where $\mathcal{M}$ represents a data analysis algorithm, $D$ and $D'$ are neighboring datasets differing in one record, $S$ is the set of possible outcomes, and $\varepsilon$ is the privacy parameter controlling the level of privacy protection.

\subsection{Rényi Differential Privacy}

Rényi Differential Privacy\cite{mironov_renyi_2017} is an extension of Differential Privacy that uses Renyi divergence to measure privacy guarantees. It is defined as:

\begin{equation}
	\label{eq:rdp_definition}
	\ln\left(\frac{\Pr[\mathcal{M}(D) \in S]}{\Pr[\mathcal{M}(D') \in S]}\right) \leq \frac{\varepsilon}{q - 1}
\end{equation}

where
$q$ is a non-negative parameter,
and other terms have the same meanings as in Differential Privacy.

\subsection{Generative Adversarial Network}

A Generative Adversarial Network is a framework for training generative models. It consists of two neural networks, a generator ($G$) and a discriminator ($D$), engaged in a two-player minimax game. The GAN objective is defined as:

\begin{equation}
	\begin{aligned}
		\min_G \max_D V(D, G) = &\mathbb{E}_{x \sim p_{\text{data}}(x)}[\log D(x)] \\
		&+ \mathbb{E}_{z \sim p_z(z)}[\log(1 - D(G(z)))]
	\end{aligned}
\end{equation}

where
$p_{\text{data}}(x)$ is the distribution of real data,
$p_z(z)$ is the distribution of random noise,
$G(z)$ represents the generated data,
and $D(x)$ is the discriminator's output representing the probability that $x$ is real.

\section{Proposed Method}

\subsection{Injection of Differential Privacy Noise in Generative Adversarial Networks}

In the current mainstream research on Differential Privacy Generative Adversarial Networks (DP-GANs), notable architectures include DP-GAN \cite{xie_differentially_2018}, DP-WGAN \cite{huang_dpwgan_2022}, dp-GAN \cite{zhang_differentially_2018}, and DP-CGAN \cite{torkzadehmahani_dp-cgan_2020}. They all employ a gradient perturbation method called DP-SGD \cite{abadi_deep_2016} for updating model parameters. The distinguishing feature of this algorithm is the addition of noise to gradients derived from each sample during model operations. These gradients are then bounded within their $l_2$ norms, and the model's weights are constrained within a specified range $\left[ { - C, + C} \right]$ using a constant $C$.

While DP-SGD indeed yields a sufficient amount of data with differential privacy, it comes at a cost of training time, often several times longer than non-differentially private model training. Additionally, because the DP-SGD algorithm perturbs gradients for all parameters of the discriminator $D$, the entire training process becomes highly unstable. Therefore, a new approach is needed to address this issue, one that adds less noise while achieving similar results.

\begin{figure}[ht]
	\centering
	\includegraphics[width=0.45\textwidth]{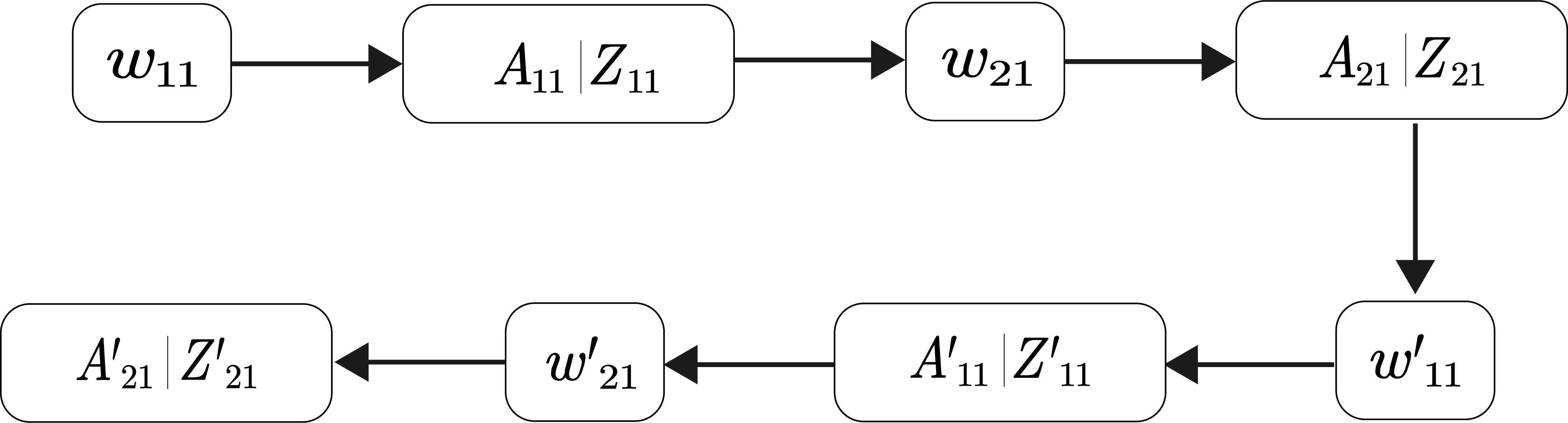}
	\caption{Assume this is a path of backpropagation in a GAN. Taking $w_{11}$ as the weight of the first neuron's input in the first layer of the generator $G$, and $A\prime_{21}\left| Z\prime_{21} \right.$ as the output of the last layer of the discriminator.}
	\label{Fig1}
\end{figure}

Our work is inspired by the work of Han et al. \cite{han_differentially_2021}. In Figure \ref{Fig1}, one of the computation flow directions for the generator $G$ and discriminator $D$ is illustrated. Here, ${w_{ij}}$ represents the weight of the ${j^{th}}$ neuron in the ${i^{th}}$ layer, ${A_i}$ denotes the output of the ${i^{th}}$ layer, where ${A_i} = {W_i}X$, and ${Z_i}$ represents the output after passing through the activation layer for ${A_i}$. This process can be expressed using the chain rule:

\begin{equation}
	\frac{{\partial E}}{{\partial {w_{11}}}} = \frac{{\partial E}}{{\partial {Z_{21}}^\prime }}\frac{{\partial {Z_{21}}^\prime }}{{\partial {A_{21}}^\prime }}\frac{{\partial {A_{21}}^\prime }}{{\partial {w_{21}}^\prime }} \cdots \frac{{\partial {Z_{11}}}}{{\partial {A_{11}}}}\frac{{\partial {A_{11}}}}{{\partial {w_{11}}}} +  \cdots
	\label{e3.1.1}
\end{equation}

Here, $E$ can be represented as the loss of the discriminator $D$.

From Equation \ref{e3.1.1}, it is clear that to obtain gradients for the farthest parameters, many terms need to be multiplied using the chain rule. For training the discriminator $D, $ since it is the module most directly exposed to sensitive data, if the gradients of the first layer of the discriminator $D$ model are not perturbed, even if gradients are perturbed for subsequent layers, it cannot guarantee that the final result will possess differential privacy properties. Taking the output $A\prime_1$ of the first intermediate layer of the discriminator as an example, $A\prime_1=W\prime_1X$, where $X$ is sensitive data without any noise perturbation. The gradients with respect to parameters $W\prime_1$, denoted as $\partial {W\prime_1}$, will have $X$ as a coefficient. This would compromise the intention of providing differential privacy to sensitive data. Hence, the DP-SGD optimization algorithm is used to perturb gradients for all parameters of the Generative Adversarial Network (GAN).

However, when considering the training process of the generator $G$, it indeed can be carried out without accessing any sensitive data. In this process, operators sample a random sample $z$ from an arbitrary distribution and input it into the generator $G$, allowing the generator $G$ to map the random sample $z$ to the target data domain. This process does not encounter the issues mentioned earlier in training the discriminator $D$, as the randomly sampled $z$ from the distribution has no practical meaning and does not involve sensitive data. Thus, the training process of the generator $G$ can, to some extent, safeguard the security of sensitive data.

\begin{figure}
	\centering
	\includegraphics[width=0.45\textwidth,scale=1.0]{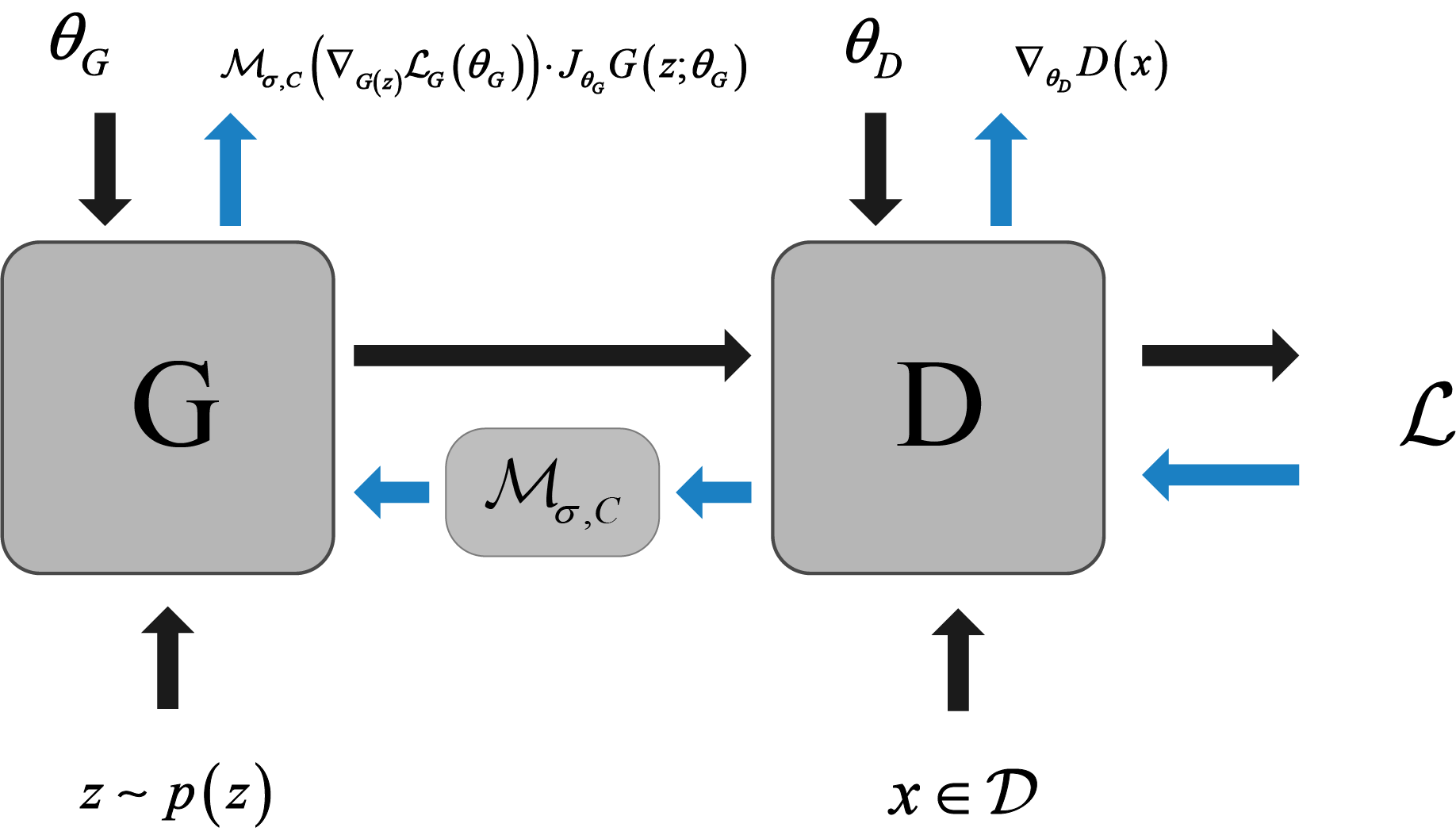}
	\caption{The GS-WGAN architecture proposed by Chen et al. \cite{chen_gs-wgan_2021}. The black arrows represent the forward propagation path, while the blue arrows indicate the backward gradient propagation path. ${\mathcal{M}_{\sigma ,C}}$ is the gradient sanitization module they designed.}
	\label{Fig2}
\end{figure}

This viewpoint is similar to the design philosophy of Chen et al. \cite{chen_gs-wgan_2021}, but they did not provide detailed descriptions. In Chen et al.'s work, their proposed GS-WGAN training architecture, as shown in Figure \ref{Fig2}, employs non-differential privacy when training the discriminator $D$, which is the traditional generative adversarial network training method. However, during the training of the generator $G$, they perform gradient sanitization on the gradients flowing from the discriminator $D$ to the generator $G$. The specific operation is calculated as follows:

\begin{equation}
	\hat{g}_G=\mathcal{M}_{\sigma ,C}\left( \nabla _{G\left( z \right)}\mathcal{L}_G\left( \theta _G \right) \right) \cdot J_{\theta _G}G\left( z;\theta _G \right)
\end{equation}

Here, $J$ represents the Jacobian matrix, $\mathcal{L}$ denotes the loss function of the generative adversarial network, and ${\theta_G}$ represents all parameters of the generator $G$. This operation eliminates the need to apply gradient perturbation to all parameters of the discriminator $D$ while ensuring the preservation of differential privacy.

From the above discussion, it is evident that when training the discriminator $D$, there is no need to perturb gradients for all parameters. Instead, it is possible to choose to perturb the gradients flowing from the discriminator $D$ to the generator $G$ during the training of the generator $G$. The advantage of doing so is to reduce the added noise and enable the discriminator $D$ to more accurately guide the generation process of the generator $G$.

To achieve this, we employ an algorithm called DP-HOOK for differential privacy noise injection. DP-HOOK is an algorithm based on the implementation of 'hook functions.' During the model initialization phase, a hook function is registered for a specific layer of the model. Subsequently, during the model training process, this hook function is triggered through backpropagation, thereby allowing manipulation of the gradient flow.

\begin{breakablealgorithm}
	\caption{DP-HOOK}
	\label{DP-SAGAN：HOOK}
	\begin{algorithmic}[1]
		\REQUIRE The input gradient $g_{in}$ during backward gradient propagation in the model layer, the $l_2$ norm bound of the gradient $C$, the batch size $B$, and the noise scale $\sigma $
		
		\STATE \textbf{$l_2$ normed input gradient}
		\STATE $g_{in}^{norm}\gets \left\| g_{in} \right\| _2$
		
		\STATE \textbf{Clip}
		\STATE $C_{bound}\gets C/B$
		\STATE $C_{coef}\gets C_{bound}/\left( g_{in}^{norm}+1e-10 \right) $
		\STATE $C_{coef}\gets \min \left( C_{coef},1 \right)$
		\STATE $g_{in}\gets C_{coef}*g_{in}$
		
		\STATE \textbf{Add Noise}
		\STATE $noise\,\,\gets C_{bound}*\sigma *\mathcal{N} \left( 0,I \right) $
		\STATE $g_{in}\gets g_{in}+noise$
		
		\STATE \textbf{output}
		\STATE $g_{out}\gets g_{in}$

		\ENSURE Output the gradient $g_{out}$ output after the reverse gradient of the model layer ends.
	\end{algorithmic}
\end{breakablealgorithm}

\begin{theorem}\cite{chen_gs-wgan_2021}
	Regarding the number of iterations for each generator update, using the DP-HOOK algorithm satisfies $\left( \lambda ,2B\lambda /\sigma ^2 \right) -RDP$, where $B$ represents the batch size.
\end{theorem}

\subsection{Tabular data modeling}

To enhance the accuracy of modeling tabular data, we adopted the data modeling approach of CTAB-GAN+\cite{zhao_ctab-gan_2022}. Specifically, we categorized tabular data types into four distinct types: continuous, discrete, mixed, and long-tail. The mixed type is a particularly unique data type that exhibits both continuous and discrete data characteristics simultaneously. To illustrate, consider a bank database that stores data for each customer's interactions with the bank, including a field labeled 'borrowing.' In reality, when a person chooses to engage in a 'borrowing' action, their borrowing amount can increase from $0$ to any real number, indicating the continuous nature of this data type. Conversely, when a person's borrowing amount changes from $0$, their borrowing status transitions from 'never borrowed' to 'borrowed,' demonstrating the discrete data characteristics. Therefore, the 'borrowing' data type is considered mixed numerical data. To be more specific, mixed numerical data exhibits 'singular values' in its distribution, where these 'singular values' represent a classification. The distribution between 'singular values' is continuous. Mathematically, mixed numerical data resembles a 'piecewise function'.

In terms of data modeling, we enforce the binding of a conditional vector as a label for each data type. This approach compels the model during training to update its parameters to align with the data distribution corresponding to that label. In doing so, it guides the Generative Adversarial Network (GAN) to generate data consistent with the conditional vector, thereby addressing data imbalance issues.

The specific processing steps are as follows: for each row of data in the table, every feature value within it is combined with its corresponding conditional vector. The conditional vector can take the form of a one-hot encoded label vector or another representation of feature conditions. Once the feature values are combined with the conditional vector, a row of input data will take on the format "value + conditional vector + value + conditional vector + ... + value + conditional vector”.

For mixed numerical data, it is essential to emphasize that this data type possesses both continuous and discrete data characteristics simultaneously. Concerning its continuous part, the probability density distribution can be effectively viewed as a mixture of multiple Gaussian distributions. Therefore, Gaussian Mixture Models (GMMs) can be employed for modeling and analysis. As for its discrete part, it naturally appears as a combination of various 'singular values.' For instance, in the case of 'borrowing,' '0' represents a 'singular value,' signifying the classification of whether there is borrowing or not.

\begin{figure}[ht]
	\centering
	\includegraphics[width=0.3\textwidth]{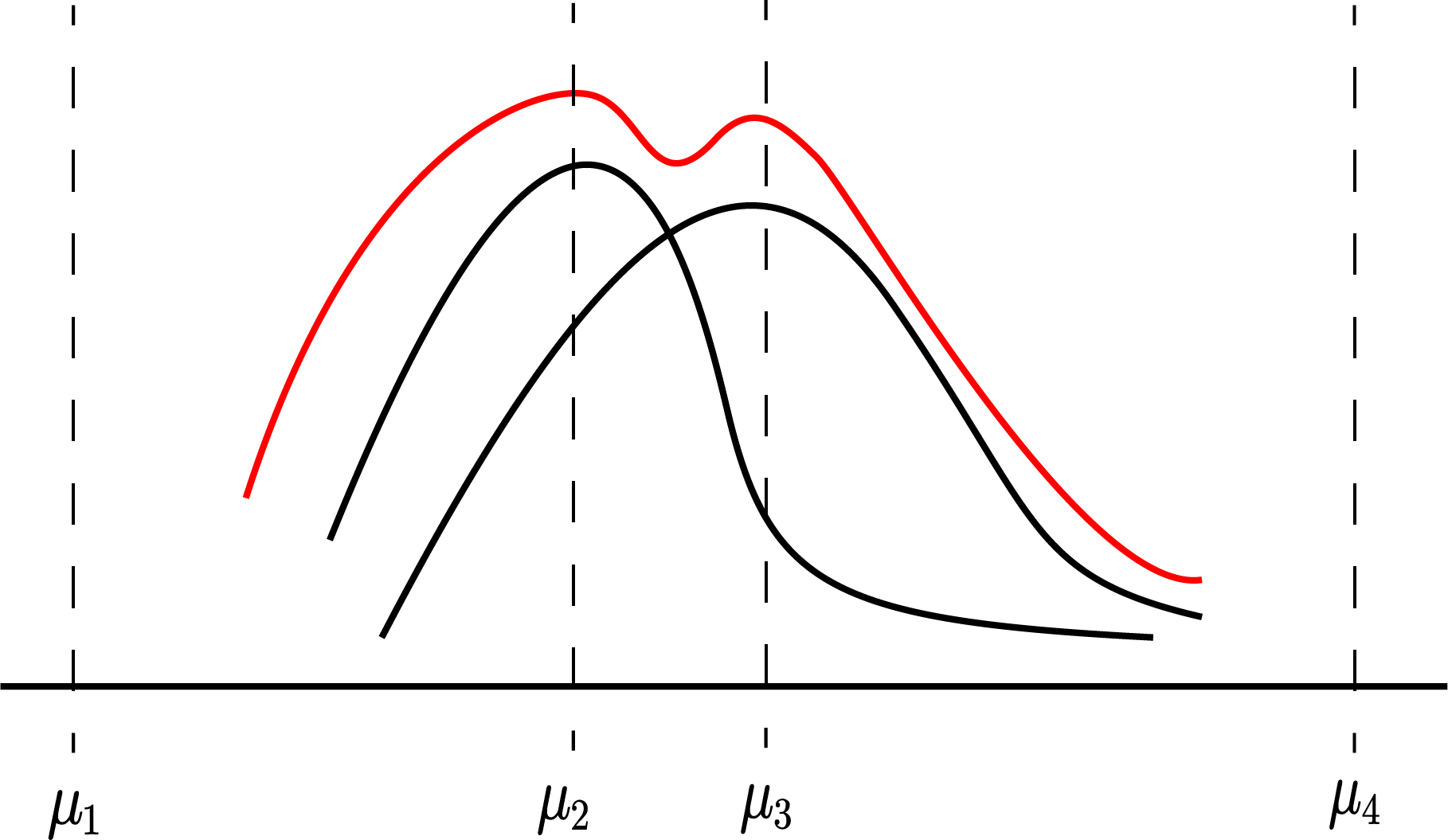}
	\caption{Example of probability density distribution for numerical data. The red line represents the final Gaussian mixture distribution. $\mu _i,i=1,2,3,4$ represents the mean value of each distribution that makes up the Gaussian mixture distribution.}
	\label{Fig3}
\end{figure}

Figure \ref{Fig3} illustrates an example of mixed numerical data. Let's assume that this data is composed of four different distributions, where each distribution is represented by its respective mean $\mu$ in the figure. From the graph, it is evident that this data comprises two discrete distributions ($\mu _1$ and $\mu _4$) and two continuous distributions ($\mu _2$ and $\mu _3$), with the continuous distributions forming the distribution of the red curve in the figure. In our work, we employ the Variational Gaussian Mixture (VGM) model\cite{bishop_pattern_2006} to analyze the continuous portion of mixed numerical data and estimate the distribution patterns it possesses. Assuming that the distribution has two patterns (as shown in the figure), i.e., $k=2$, the Gaussian mixture distribution learned by VGM is as follows:

\begin{equation}
	\mathbb{P} =\sum_{k=1}^2{\omega _k\mathcal{N} \left( \mu _k,\sigma _k \right)}
\end{equation}

Among them, $\mathbb{P}$ represents the finally learned Gaussian mixture distribution, and the parameters $\omega _k$, $\mu_k$, and $\sigma_k$ owned by the normal distribution $\mathcal{N}$ represent respectively The weight, mean, and variance of the distribution.

\begin{figure}[ht]
	\centering
	\includegraphics[width=0.3\textwidth]{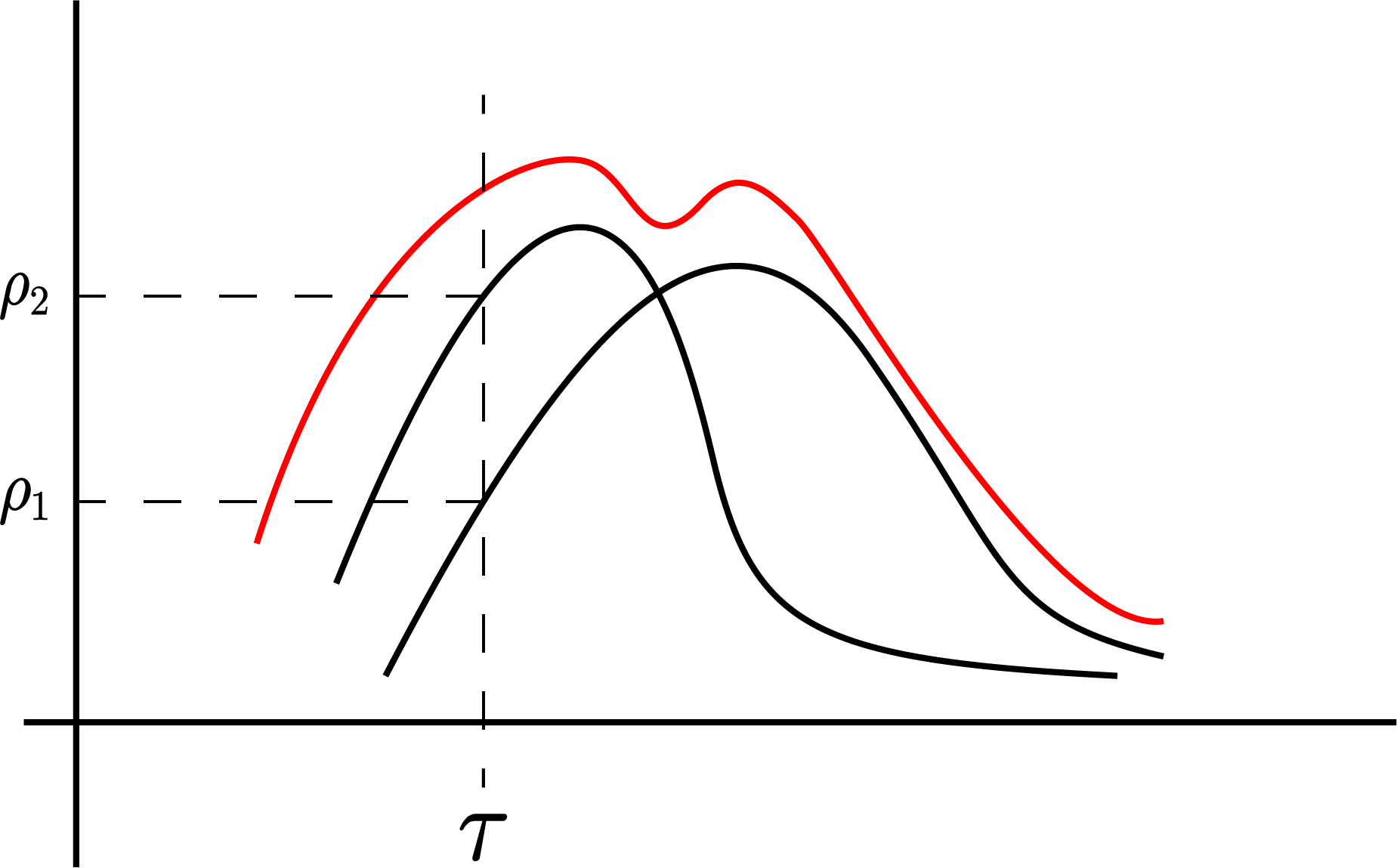}
	\caption{Randomly sample from a continuous distribution of this mixed data type, and for a single sampled value $\tau$, it has a corresponding probability density value on each different pattern.}
	\label{Fig4}
\end{figure}

Since normalizing the entire distribution as input into VGM requires normalization of each value in the distribution, in this work, we associate each randomly sampled value with the model corresponding to the highest probability density. Figure \ref{Fig4} illustrates this concept effectively. It is clear from the figure that the randomly sampled value $\tau$ corresponds to the highest probability density, which is $\rho _2$. Therefore, this probability density mode will be involved in the normalization calculation. The specific formula is as follows:

\begin{equation}\label{DP-SACTGAN：equ1}
	\alpha =\frac{\tau -\mu _2}{4\sigma _2}
\end{equation}

The symbol $\alpha$ represents the value after the transformation of $\tau$. After the numeral system conversion is completed, according to the data coding rules, a label corresponding to the numerical value will be appended to it. The labels here will be encoded as one-hot vectors based on the number of modes in the overall distribution. Taking Figure \ref{Fig4} as an example, the conditional vector corresponding to $\alpha$ is $\beta =\left[ 0,1,0,0 \right]$, indicating that the value has been normalized using mode 2.

Considering the classification characteristics of this mixed numerical type data, it does not need to be placed into the VGM for mode estimation. However, to ensure the integrity of the data format, when sampling the classification part of the data, the format conversion is still performed according to the predetermined data coding format. For example, if $\mu_1$ is sampled in Figure \ref{Fig4}, the conditional vector after conversion for this value is $\beta =\left[ 1,0,0,0 \right] $, indicating that the value has been normalized using mode 1.

\subsection{DP-SACTGAN}

\begin{figure}[ht]
	\centering
	\includegraphics[width=0.45\textwidth]{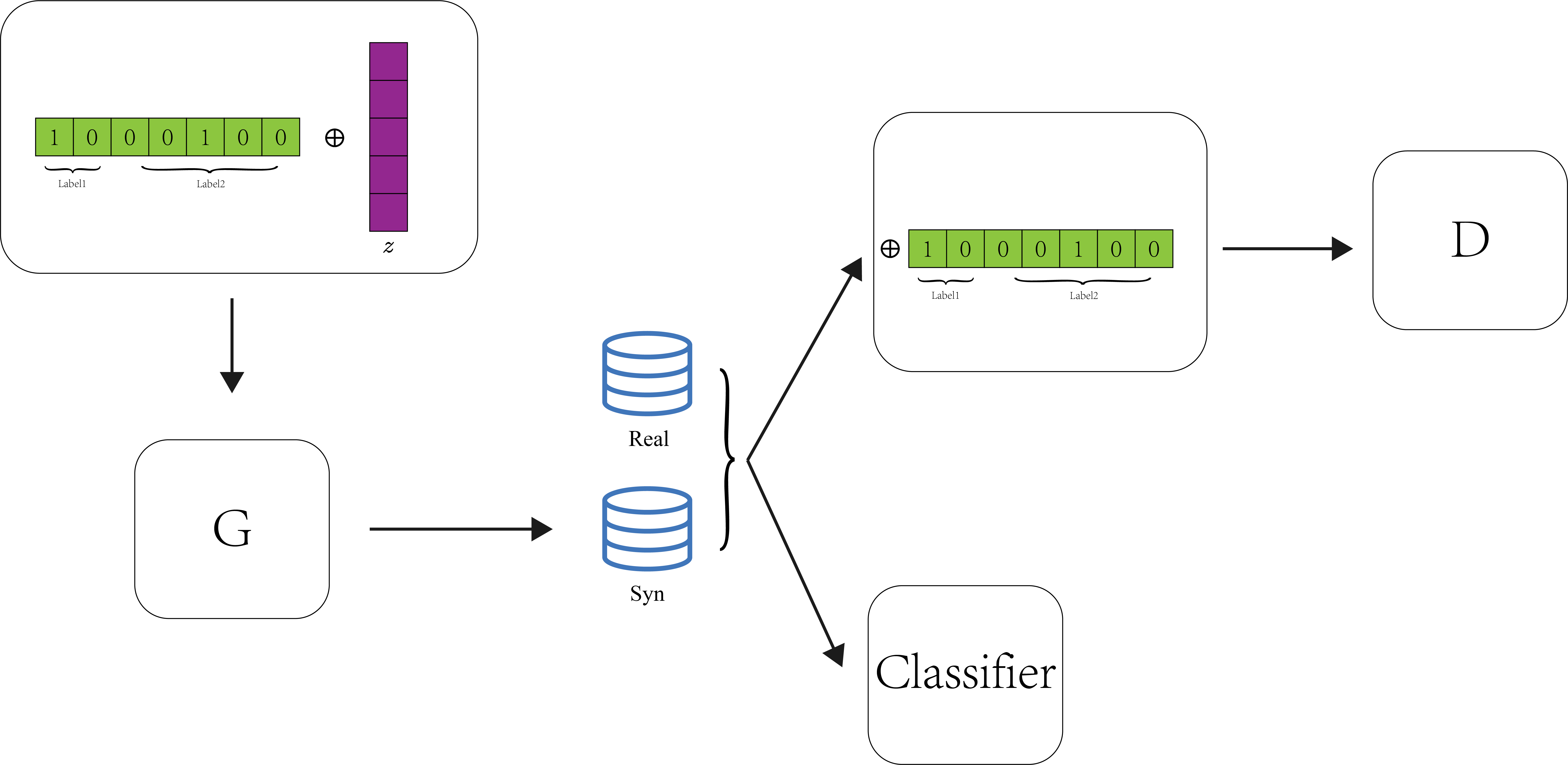}
	\caption{The training process of DP-SACTGAN.}
	\label{DP-SACTGAN Train}
\end{figure}

\begin{figure}[ht]
	\centering
	\includegraphics[width=0.45\textwidth]{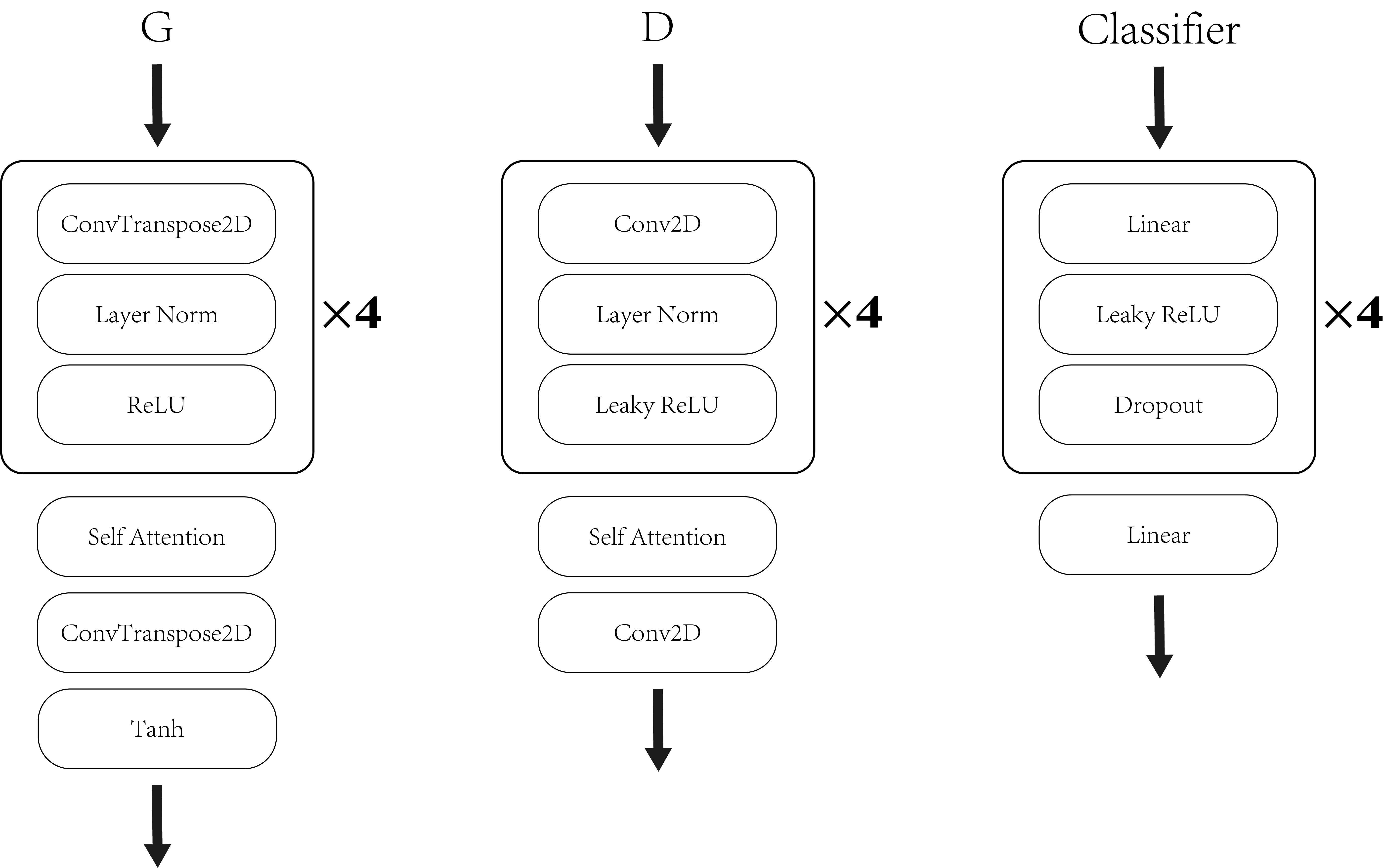}
	\caption{The complete architecture of each module of DP-SACTGAN}
	\label{Fig5}
\end{figure}

Figure \ref{Fig5} presents the comprehensive architecture of DP-SACTGAN. In general, each data point in tabular data is relatively independent, requiring individual computation and processing. Therefore, we have opted for layer normalization as the choice for normalization layers.

Additionally, the auxiliary classifier module introduced in DP-SACTGAN is essentially a multi-layer perceptron (MLP) structure designed to provide greater information support for model training. This auxiliary classifier consists of a series of fully connected layers, comprising four intermediate layers and one output prediction classification layer. During the training process, the predictions from the auxiliary classifier are used as the basis for updating the generator, effectively introducing an additional form of supervision signal to the generator.

Figure \ref{DP-SACTGAN Train} depicts the entire training process of DP-SACTGAN.

\section{Experiments}

In this research task, Ubuntu 20.04 operating system was employed as the experimental environment. The processing unit used for the experiments was an Intel® i7-9700k. To expedite the model training process, a GeForce NVIDIA® 2080Ti 11GB graphics card was utilized. On the software front, PyTorch® version 1.12.1 was employed as the deep learning framework, with GPU acceleration achieved through CUDA version 11.4.

To intuitively showcase the performance of DP-SACTGAN, we selected the Adult dataset \cite{kohavi_scaling_1996} and the King dataset \cite{noauthor_house_nodate}. For model selection, we chose CTAB-GAN+ \cite{zhao_ctab-gan_2022}, CTGAN \cite{xu_modeling_2019}, DP-GAN \cite{xie_differentially_2018}, and PATE-GAN \cite{jordon_pate-gan_2018} as comparative models for the experimental section of this paper.

\subsection{Data utility analysis}

\begin{table}[htbp!]
	\centering
	\footnotesize
	\begin{tabular}{lrrr}
		\toprule
		\multicolumn{4}{c}{Adult LR $\varepsilon$=1} \\
		\midrule
		Models   & \multicolumn{1}{l}{ACC} & \multicolumn{1}{l}{AUC} & \multicolumn{1}{l}{F1} \\
		\midrule
		DP-SACTGAN & \textbf{4.084348} & \textbf{0.026138} & \textbf{0.02508} \\
		CTAB-GAN+ & 5.046576 & 0.056005 & 0.08038 \\
		CTGAN & 5.282015 & 0.361938 & 0.20729 \\
		DP-GAN & 5.077285 & 0.482824 & 0.22059 \\
		PATE-GAN & 46.852288 & 0.389049 & 0.31661 \\
		\bottomrule
	\end{tabular}%
	\caption{Data utility on logistic regression (LR) for the synthetic Adult dataset of $\varepsilon=1$.}
	\label{table1}
\end{table}

\begin{table}[htbp!]
	\centering
	\footnotesize
	\begin{tabular}{lrrr}
		\toprule
		\multicolumn{4}{c}{King Ridge $\varepsilon$=1} \\
		\midrule
		Models   & \multicolumn{1}{l}{MAE} & \multicolumn{1}{l}{EVS} & \multicolumn{1}{l}{R2} \\
		\midrule
		DP-SACTGAN & \textbf{0.011382} & \textbf{0.165298} & \textbf{0.2143} \\
		CTAB-GAN+ & 0.041995 & 0.208086 & 0.22156 \\
		CTGAN & 0.968496 & 0.701021 & 1.6096 \\
		DP-GAN & 10.983962 & 9.143848 & 177.077 \\
		PATE-GAN & 3.656045 & 2.85261 & 19.1745 \\
		\bottomrule
	\end{tabular}%
	\caption{Data utility on Ridge regression (Ridge) for the synthetic King dataset of $\varepsilon=1$.}
	\label{table2}
\end{table}

It is evident from Table \ref{table1} and \ref{table2} that DP-SACTGAN outperforms other models in all three evaluation metrics on all datasets when it comes to downstream classification. In some cases, CTAB-GAN+ and CTAB-GAN exhibit better performance than DP-SACTGAN.

DP-GAN, PATE-GAN, PATE-CTGAN, and PATE-CTDRAGAN, which do not employ precise data modeling and have been trained with differential privacy methods, generally yield subpar results on the generated data.

\subsection{Statistical evaluation of data}

For the evaluation of all continuous and mixed columns, the Wasserstein Distance (WD) is used. For all discrete columns, the Jensen-Shannon Divergence (JSD) is employed. In all cases, the evaluation results are presented using a difference-based approach.

\begin{table}[htbp!]
	\centering
	\footnotesize
	\begin{tabular}{lrrr}
		\toprule
		\multicolumn{4}{c}{Adult $\varepsilon$=1} \\
		\midrule
		Models   & \multicolumn{1}{l}{WD} & \multicolumn{1}{l}{JSD} & \multicolumn{1}{l}{Diff Cor} \\
		\midrule
		DP-SACTGAN & \textbf{0.014889} & 0.101352 & \textbf{0.84923} \\
		CTAB-GAN+ & 0.015372 & \textbf{0.080305} & 1.25747 \\
		CTGAN & 0.199956 & 0.723736 & 2.39017 \\
		DP-GAN & 0.151921 & 0.409907 & 2.29194 \\
		PATE-GAN & 0.392529 & 0.51769 & 2.12781 \\
		\bottomrule
	\end{tabular}%
	\caption{Statistical evaluation of generated Adult tabular data for all baseline models with $\varepsilon =1$}
	\label{table3}
\end{table}

\begin{table}[htbp!]
	\centering
	\footnotesize
	\begin{tabular}{lrrr}
		\toprule
		\multicolumn{4}{c}{King $\varepsilon$=1} \\
		\midrule
		Models   & \multicolumn{1}{l}{WD} & \multicolumn{1}{l}{JSD} & \multicolumn{1}{l}{Diff Cor} \\
		\midrule
		DP-SACTGAN & \textbf{0.029829} & \textbf{0.102852} & \textbf{2.63088} \\
		CTAB-GAN+ & 0.031369 & 0.132738 & 3.31095 \\
		CTGAN & 0.218678 & 0.74488 & 5.78749 \\
		DP-GAN & 0.273659 & 0.832943 & 5.11406 \\
		PATE-GAN & 0.353246 & 0.728053 & 5.33376 \\
		\bottomrule
	\end{tabular}%
	\caption{Statistical evaluation of generated King tabular data for all baseline models with $\varepsilon =1$}
	\label{table4}
\end{table}

From Tables \ref{table3} and \ref{table4}, it is evident that DP-SACTGAN generally outperforms the baseline models in most of the evaluation metrics. DP-SACTGAN's WD distance evaluation results are the best for the King dataset at $\varepsilon=1$ privacy strength, but in other tasks, such as the Adult dataset at $\varepsilon=1$ privacy strength, DP-SACTGAN's JSD distance evaluation results do not surpass those of CTAB-GAN+. This suggests that DP-SACTGAN, with its self-attention mechanism, is more adept at fitting continuous distributions rather than discrete ones.

However, DP-SACTGAN excels in all Diff Cor metric evaluations compared to other models, demonstrating the effectiveness of the self-attention mechanism in providing global information integration.

\subsection{Member inference attack experiment}

\begin{figure}[ht]
	\centering
	\includegraphics[width=0.45\textwidth]{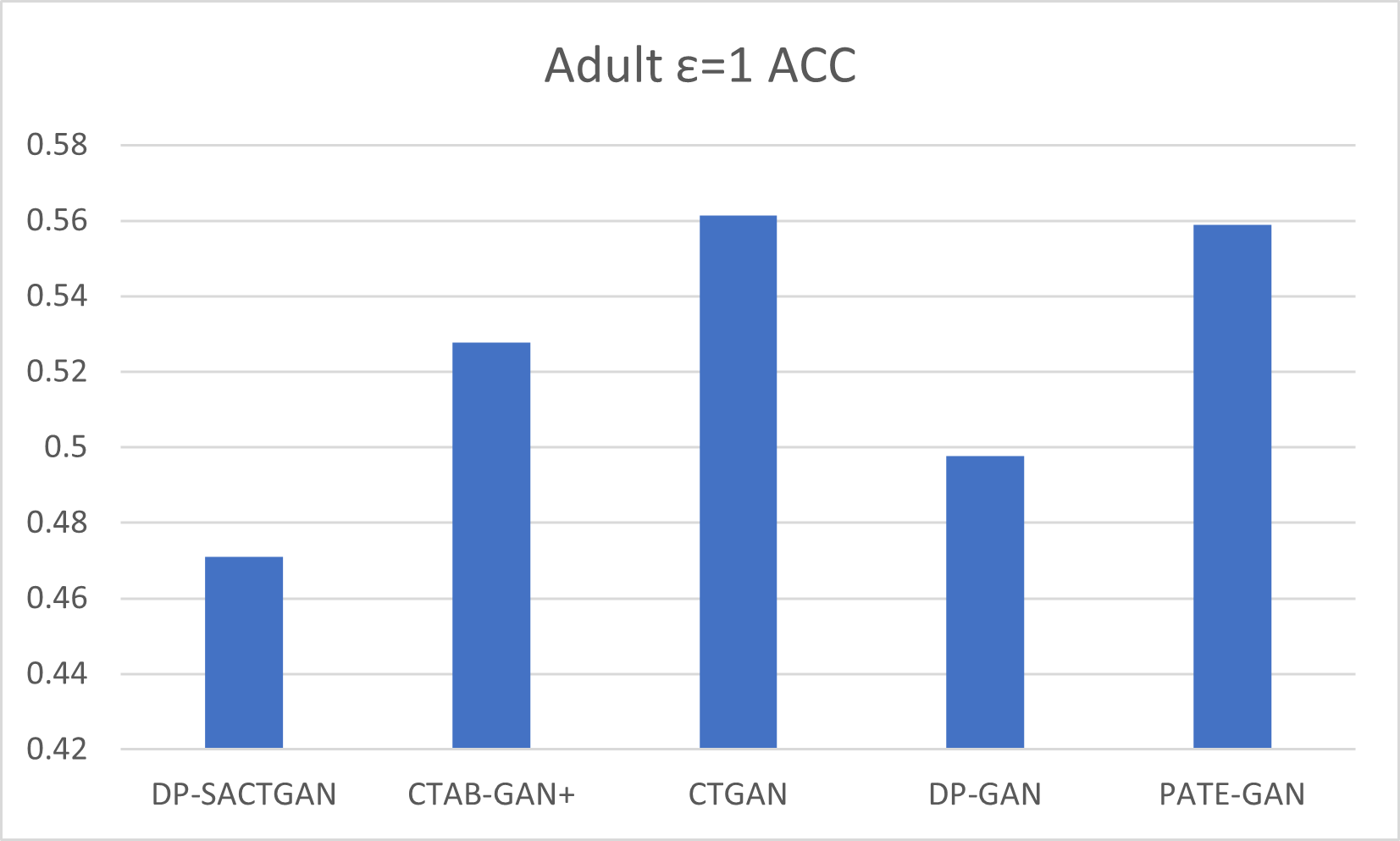}
	\caption{Accuracy of all models under the attacker's attack on the $\varepsilon=1$ Adult synthetic data set}
	\label{Fig6}
\end{figure}

\begin{figure}[ht]
	\centering
	\includegraphics[width=0.45\textwidth]{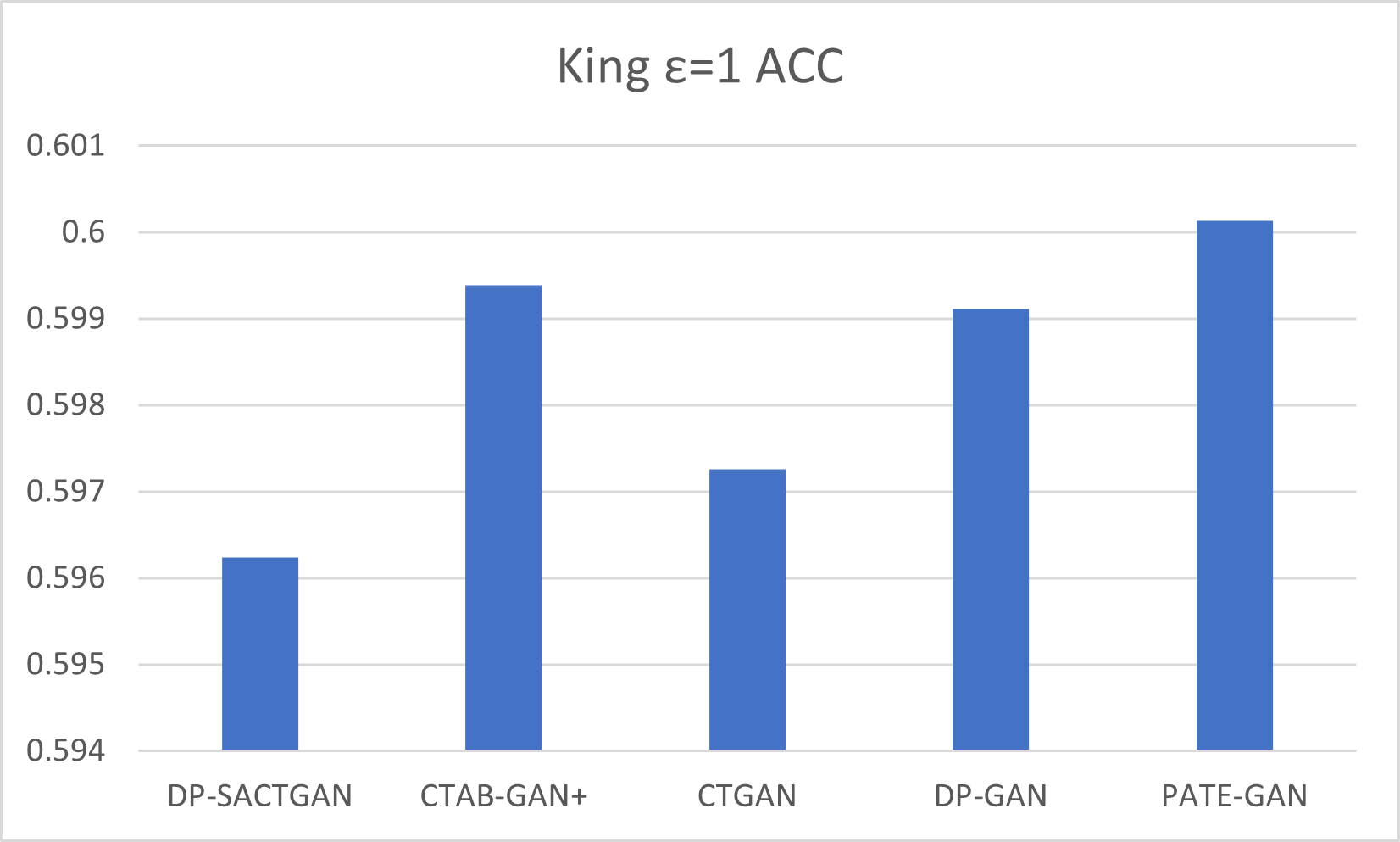}
	\caption{Accuracy of all models under the attacker's attack on the $\varepsilon=1$ King synthetic data set}
	\label{Fig7}
\end{figure}

From Figure \ref{Fig6} and Figure \ref{Fig7}, it can be observed that at a privacy strength of $\varepsilon=1$, the DP-HOOK noise addition method used by DP-SACTGAN performs roughly on par with, or even outperforms, models trained using PATE and DP-SGD methods in the membership inference attack. This demonstrates the effectiveness of our approach.

\section{Conclusion}

This paper delves deep into the architectural properties of Generative Adversarial Networks (GANs) and discovers that during the training process using a generator, there is no need to access the privacy dataset at any point. Therefore, it becomes unnecessary to add noise on a per-sample basis since the generator's input inherently consists of random noise. Furthermore, in the process of training the generator, adding noise to the gradient flow passed from the discriminator to the generator is sufficient to ensure differential privacy, providing the generator with more accurate guidance without introducing excessive noise.

The proposed DP-SACTGAN in this paper offers specific modeling approaches for each column of tabular data and employs the DP-HOOK method to guarantee the differential privacy of generated data. Experimental results show that our method performs comparably to classical differential privacy methods and even excels in terms of data utility.

\bibliographystyle{plain}
\bibliography{main}

\end{document}